%% file: main.tex
\documentclass[conference]{IEEEtran}
\usepackage{times}

\usepackage[numbers]{natbib}
\usepackage{multicol}
\usepackage[usenames,dvipsnames,table]{xcolor}
\usepackage[bookmarks=true]{hyperref}
\hypersetup{
    colorlinks=true,
    linkcolor=orange,
    filecolor=magenta,      
    urlcolor=orange,
    citecolor=orange,
}
\usepackage{pdfx}
\usepackage{booktabs}
\usepackage{xspace}
\usepackage{amsmath}
\usepackage{caption}
\usepackage[capitalize]{cleveref}
\usepackage{graphicx}
\usepackage{inconsolata}
\usepackage[all]{hypcap}

\input{macros}

\begin{document}

\title{Action-Free Reasoning for Policy Generalization}


\author{\authorblockN{\textbf{Jaden Clark, Suvir Mirchandani, Dorsa Sadigh, Suneel Belkhale}}
\authorblockA{Department of Computer Science, Stanford University\\
\renewcommand{\ttdefault}{pcr}
\texttt{\{jvclark, smirchan, dorsa, belkhale\}@stanford.edu}
}}


%

\makeatletter
\let\@oldmaketitle\@maketitle
\renewcommand{\@maketitle}{\@oldmaketitle
  \centering
  \includegraphics[width=0.98\linewidth]{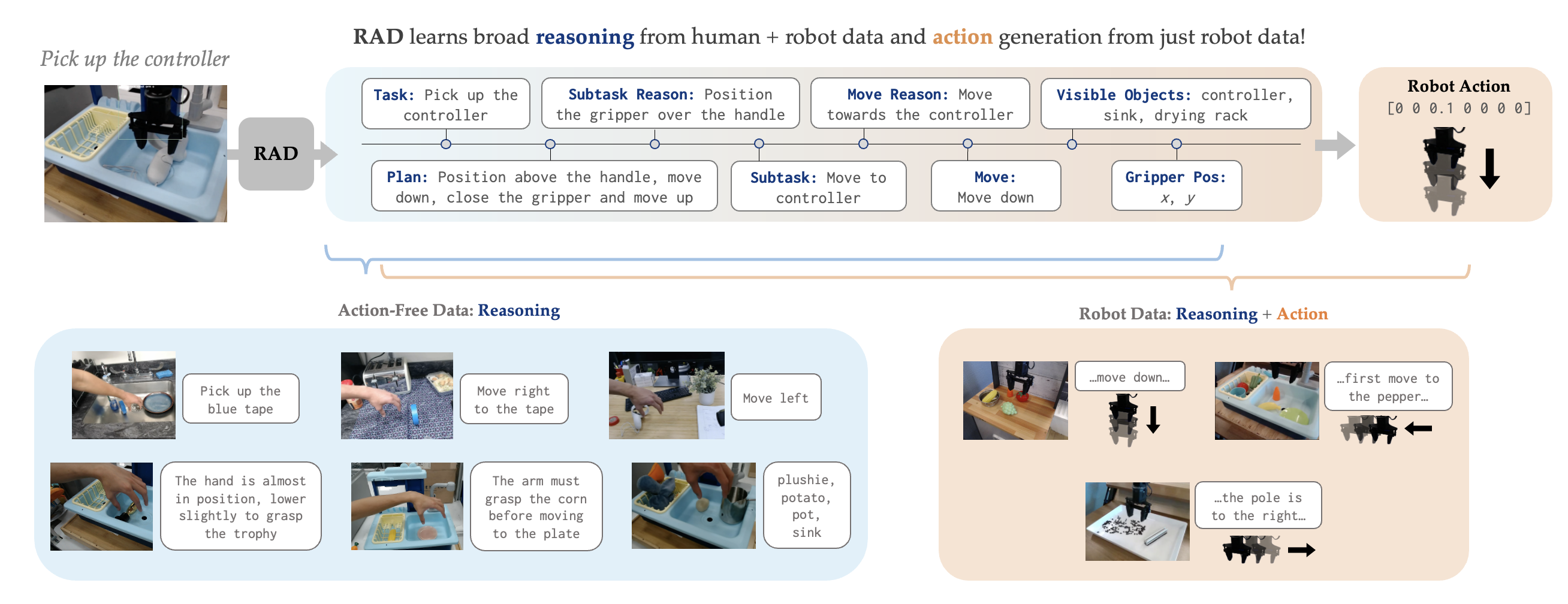}

    \captionof{figure}{\ACRO learns from both human and robot data through chain-of-thought reasoning. \ACRO learns how to reason through high level task plans, subtasks, and movements from human data, and how to map reasonings to action from robot data. This enables \ACRO models to generalize to tasks unseen in both human and robot data.}
    \label{fig:front}
    }%
\makeatother

\maketitle

\begin{abstract}
\input{sections/00_abstract.tex}
\end{abstract}

\IEEEpeerreviewmaketitle

\section{Introduction}
\label{sec:introduction}
\input{sections/01_introduction}
\section{Related Work}
\label{sec:related_work}
\input{sections/02_related_work}

\section{\Method}
\label{sec:method}
\input{sections/03_method}

\section{Experiments}
\label{sec:experiments}

\input{sections/04_experiments}

\section{Discussion}
\label{sec:discussion}
\input{sections/05_discussion}

\section{Acknowledgements}
We thank Priya Sundaresan for and Juntao Ren for code for setting up HaMeR tracking. We thank Jonathan Yang, Satvik Sharma, Rahul Chand, and Priya Sundaresan for paper writing support.

This work was supported by NSF \#2006388 and \#2125511, DARPA YFA Grant \#W911NF2210214, ONR N00014-21-1-2298, Toyota Research Institute, and DARPA TIAMAT project.



\bibliographystyle{plainnat}
\bibliography{references}
\newpage
\section*{Appendix}
\label{sec:appendix}

\input{sections/06_appendix}

\end{document}

%% file: macros.tex
\newcommand{\Method}{Reasoning through Action-free Data\xspace}
\newcommand{\ACRO}{RAD\xspace}

\newcommand \instr[1]{\textit{#1}}

%% file: sections/00_abstract.tex
End-to-end imitation learning offers a promising approach for training robot policies. However, generalizing to new settings—such as unseen scenes, tasks, and object instances—remains a significant challenge. Although large-scale robot demonstration datasets have shown potential for inducing generalization, they are resource-intensive to scale. In contrast, human video data is abundant and diverse, presenting an attractive alternative. Yet, these human-video datasets lack action labels, complicating their use in imitation learning. Existing methods attempt to extract grounded action representations (e.g., hand poses), but resulting policies struggle to bridge the embodiment gap between human and robot actions.
We propose an alternative approach: leveraging language-based reasoning from human videos - essential for guiding robot actions - to train generalizable robot policies. Building on recent advances in reasoning-based policy architectures, we introduce Reasoning through Action-free Data (RAD). RAD learns from both robot demonstration data (with reasoning and action labels) and action-free human video data (with only reasoning labels). The robot data teaches the model to map reasoning to low-level actions, while the action-free data enhances reasoning capabilities. Additionally, we will release a new dataset of 3,377 human-hand demonstrations compatible with the Bridge V2 benchmark. This dataset includes chain-of-thought reasoning annotations and hand-tracking data to help facilitate future work on reasoning-driven robot learning.
Our experiments demonstrate that RAD enables effective transfer across the embodiment gap, allowing robots to perform tasks seen only in action-free data. Furthermore, scaling up action-free reasoning data significantly improves policy performance and generalization to novel tasks. These results highlight the promise of reasoning-driven learning from action-free datasets for advancing generalizable robot control. 
Website: \href{https://rad-generalization.github.io}{here}.

%% file: sections/01_introduction.tex
Training visuomotor policies via imitation learning is an appealing paradigm for robot control. However, an outstanding challenge for current end-to-end learning methods is to generalize to new settings beyond their training data, such as new scenes, new task instructions, and new object instances. For example, if a robot learns to pick up a video game controller in a lab setting, but encounters the same controller in an office, it should still use its prior knowledge to bridge the gap between different environments. 
The ability to generalize to these types of novel scenarios is essential for making learning-based policies useful in practice, as the real-world often presents diverse and unpredictable scenarios. 

One approach to achieving generalizable policies is to collect large-scale robot demonstration datasets across tasks and embodiments and train expressive multi-task policies on them \cite{khazatsky2024droid, oxe2024, kim2024openvla, team2024octo}. While there are promising signs of scaling up datasets being the solution, we simply have not reached the scale needed for comprehensive generalization, and one might argue that collecting data at such scale is practically infeasible.

On the other hand, many see tapping into human video datasets, consisting of humans directly performing tasks as opposed to collecting robot data, as the answer \cite{ye2024latent, wang2023mimicplay, bharadhwaj2024gen2act}. This data is cheap to collect and already present at scale in Internet datasets. However, human videos lack action labels, making supervised learning methods like imitation learning very difficult. Some works tackle this challenge by extracting \emph{grounded action-like} representations from video as labels for imitation learning, for example hand poses or object affordances~\cite{bharadhwaj2024track2act, ren2025motion, lepertshadow, xu2023xskill}. However, extracting grounded actions from human videos often makes assumptions about the scene and the embodiment gap (e.g., how the hand pose maps to the robot action or relying on paired human and robot data) which can limit their usefulness in practice. 

Instead of extracting grounded actions from videos and the restrictive assumptions that come with it, we ask: 
\emph{is there any other behavioral information -- that still directly influences robot actions -- that we can extract from human videos, and more generally action-free data?}
Our insight is that human videos contain vast amounts of \emph{higher-level reasoning} that guide robot action prediction, and this reasoning information can be captured via language. For example, if the task is to pick up a cup, a human hand might move towards the cup, then grasp the cup, and then lift the cup. 
Whereas prior work might learn a policy that outputs changes in the hand pose and hope this transfers to the robot, we can instead learn to predict this detailed reasoning itself from human videos, which is shared across many embodiments.
Policies that autoregressively predict each stage of reasoning -- predicting from high-level language down to grounded robot actions -- show steerable behaviors that improve performance and generalization~\cite{zawalski2024robotic, belkhale2024rt}. While these prior works often rely on reasoning over robot trajectories, our key idea is to instead extract such reasoning from \emph{action-free} human videos -- significantly scaling up data that informs robot actions.

We introduce our method, \Method (\ACRO), a robot policy that leverages reasoning traces extracted from action-free data. \ACRO trains a large transformer model on a mixture of robot demonstration data with both reasoning and robot action labels, and action-free (human video) data labeled with \emph{just} reasoning. The robot data teaches the model to autoregressively go from reasoning to low-level actions, while the action-free data augments the reasoning knowledge, thus boosting the reasoning capabilities of the model. We label reasoning traces by leveraging pretrained vision-language models such as Gemini with hindsight knowledge as done in prior work~\cite{zawalski2024robotic}. 

We experimentally validate that learning from action-free reasoning data transfers well across the embodiment gap -- showing 20\% better performance on tasks only seen in the action-free data over models not finetuned with \ACRO. Additionally, we demonstrate that having larger amounts of action-free reasoning data improves the capacity of the model to generalize in language space to completely unseen tasks with \ACRO outperforming baselines by 15\% on generalization tasks (that have never been seen in robot or human data).

%% file: sections/02_related_work.tex
In this section, we situate our work among prior work on the use of language as a representation of low-level actions in robot learning, vision-language-action models (VLAs) as a recipe for language-conditioned robot policies, and approaches that leverage human videos for robot learning.


\noindent \textbf{Language as an Action Representation.}
Language is commonly used as a high-level representation in imitation learning, either for conditioning multi-task policies on specific instructions  ~\citep{stepputtis2020language, jang2022bc, rt12022, rt22023, kim2024openvla}, or as a way to decompose high-level, long-horizon instructions into lower-level subtask instructions \citep{brohan2023do, huang2023inner, shi2024yell}. More recently, several works have studied the role of more fine-grained language such as ``language motions'' as intermediate representations to predict~\cite{belkhale2024rt} or explicitly reason over language as well as other visually-grounded features such as bounding boxes as a way of guiding large pretrained policies~\cite{zawalski2024robotic}. In contrast to prior works which use language as a goal representation, we explore how reasoning in language can be used as an action representation for human video data in addition to robot data.

\noindent \textbf{Vision Language Action Models.}
Recent works have explored the use of pre-trained Vision-Language Models (VLMs) as backbones for Vision-Language Action Models (VLAs) which directly predict low-level robot actions. For example, RT-2-X \cite{oxe2024} fine-tunes the 55B-parameter PaLI-X VLM \cite{chen2024palix} on the Open-X Embodiment dataset \cite{oxe2024}, and OpenVLA \cite{kim2024openvla} uses a 7B-parameter Llama 2 LLM backbone with a vision encoder based on DINOv2 \cite{oquab2023dinov2} and SigLIP \cite{zhai2023sigmoid}. The promise of VLAs for manipulation is to build off of generalization of VLMs which have been trained on Internet-scale vision-language data. An additional way to achieve transfer of VLM capabilities to VLAs is to take advantage of their textual reasoning abilities. For example, Embodied Chain of Thought (ECoT) uses multiple steps of reasoning prior to predicting robot actions by training on synthetic reasoning data \cite{zawalski2024robotic}.

\noindent \textbf{Learning from Human Video.}
A large number of prior works in imitation learning for robotics focus on learning from demonstrations collected via teleoperation by expert operators.
This method of collecting data is costly, so a number of prior works have investigated ways to leverage existing data sources of human videos to improve robot policy learning --- for example, by pre-training visual representations \citep{nair2022r3m, xiao2022masked, karamcheti2023language},  learning reward functions \citep{shao2021concept2robot, chen2021learning, mandikal2022dexvip}.
However, bridging the gap between human videos and robot actions can be challenging due to embodiment differences and diversity in videos. Several works learn priors from human video datasets and/or in-domain human videos \citep{shaw2023videodex, bahl2022human, wang2023mimicplay, lepertshadow} or aligning paired/unpaired examples of human videos and robot demonstration videos \citep{sharma2019third, smith2019avid, xiong2021learning, jain2024vid2robot} or simulations \citep{qin2022dexmv}. These works are still fundamentally limited by the quantity of robot demonstrations. Another line of work leverages intermediate representations for predicting robot actions downstream, but make assumptions about the human hand behavior, which is not necessarily the same as the robot \cite{papagiannis2024r+, bharadhwaj2024track2act}. Our work goes beyond existing methods that rely on generating intermediate representations for action predictions by generating detailed reasoning steps about human video demonstrations.

%% file: sections/03_method.tex
In this section, we will first describe our problem setting and lay out our assumptions, and then we will outline our method for learning from action-free data using language reasoning chains.
As an overview, \ACRO involves two major steps. First, annotate action-free data with language reasoning (\cref{sec:method:labeling}). Second, train a reasoning-based policy on a combination of robot demonstration data with both actions and reasoning chains and action-free data with only reasoning chains (\cref{sec:method:training}).

\begin{figure*}[!ht]
    \centering
    \includegraphics[width=\linewidth]{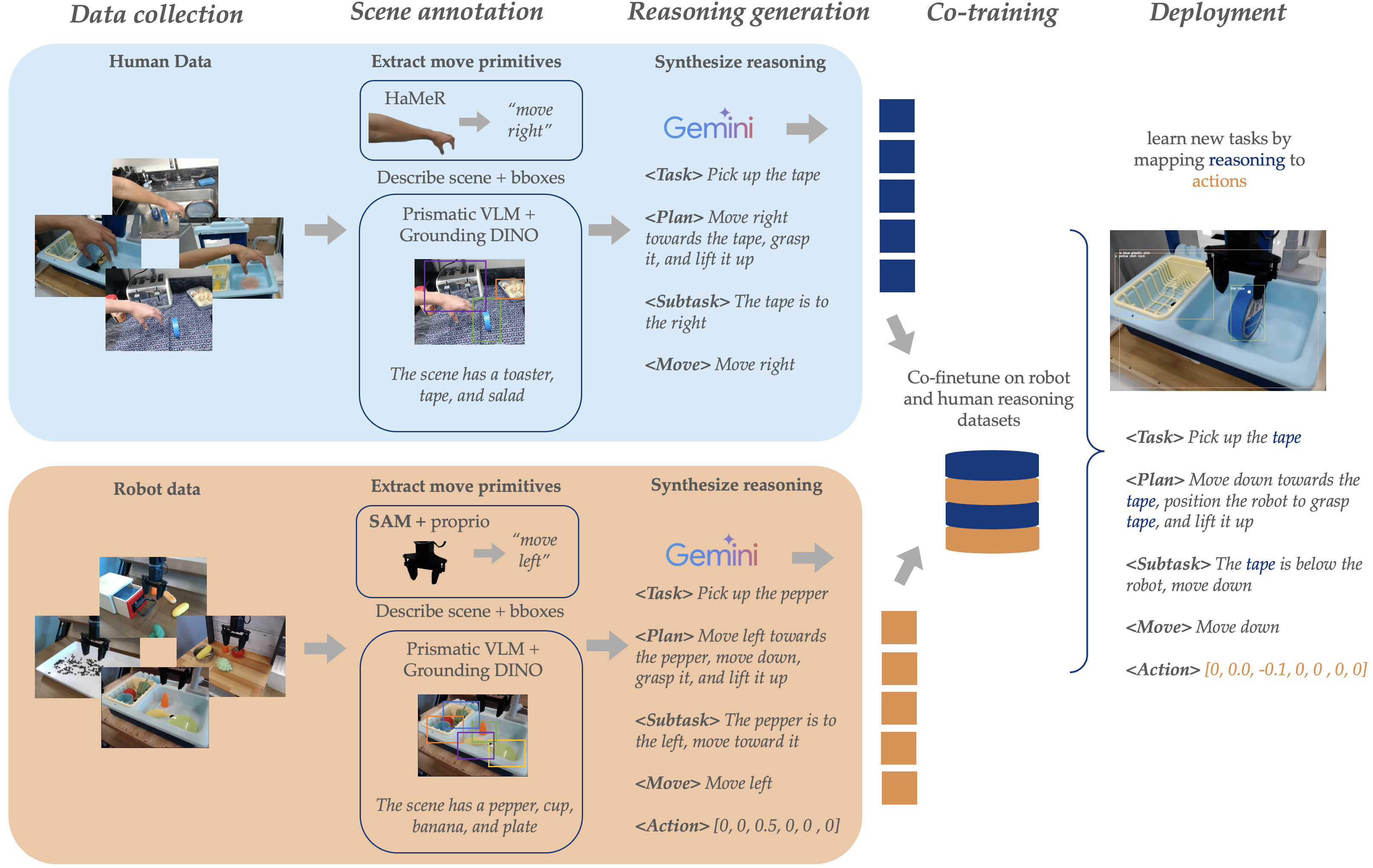}
    \caption{\ACRO generates reasonings on both human and robot data using a suite of pretrained models. Scene descriptors and object bounding boxes for both human and robot data are generated using Prismatic VLM and Grounding DINO. While SAM and proprioception can be used to generate movement primitives for robot data, \ACRO relies on HaMeR to track human hand data for primitive generation. For both data types, the scene descriptions, bounding boxes, and movement primitives (as well as actions for robot data) are synthesized by Gemini into reasoning data in natural language. These reasonings are tokenized and fed into a mixed dataset containing both human and robot data for co-finetuning.}
    \label{fig:labeling}
\end{figure*}

\subsection{Problem: Learning Reasoning in Action-free Data}

In multi-task imitation learning, we are given a dataset $\mathcal{D} = \{ (o_1, a_1, g_1),  \ldots (o_N, a_N, g_N) \}$ consisting of tuples of observations $o \in \mathcal{O}$, actions $a \in \mathcal{A}$, and task specifications $g \in \mathcal{G}$ which are often formulated in language. The objective is to learn the expert action distribution $P(a \mid o,g)$ conditioned on an observation $o$ and a task specification $g$.

We now define the objective of \textit{reasoning-based} multi-task imitation learning. We assume there exists some chain of $C$ steps of intermediate language reasoning that links an observation $o$ and action label $a$, which we denote as $(l^1, \ldots, l^C)$. We discuss how these reasoning chains are generated in \cref{sec:method:labeling}. The distribution of each reasoning step $l^j$ only depends on the preceding reasoning steps $(l^1, \dots, l^{j-1})$ as well as $o$ and $g$. The distribution of actions $a$ depends on all reasoning steps $(l^1, \ldots, l^C)$ and the observation $o$ and task $g$. We define the objective of the reasoning-based multi-task imitation learning problem as learning the expert joint reasoning and action distribution $P(a, l^1, \ldots, l^C \mid  o, g)$. In this setting, each $(o_i, a_i, g_i)$ tuple in $\mathcal{D}$ is augmented with a reasoning chain $(l_i^1, \ldots, l_i^C)$. We wish to learn a distribution $P_\theta$ parameterized by $\theta$ that  maximizes the log-likelihood of the reasoning and action data in $\mathcal{D}$:
\begin{align*}
    L(\theta)& = \sum_i^N \log P_\theta(a_i,l_i^1\dots l_i^C \mid o_i,g_i) \\
    =\ &\sum_i^N \log P_\theta(a_i \mid l_i^1\dots l_i^C,o_i,g_i) \prod_j^C P_\theta(l_i^j \mid l_i^1\dots l_i^{j-1},o_i,g_i)  \\
    =\ &\sum_i^N \log P_\theta(a_i \mid l_i^1\dots l_i^C,o_i,g_i) \\
    & + \sum_i^N \sum_j^C \log P_\theta(l_i^j \mid l_i^1\dots l_i^{j-1},o_i,g_i) \\
    =\  &L_{\texttt{action}}(\theta) + L_{\texttt{reasoning}}(\theta)
\end{align*}

Our key insight in \ACRO is that \textit{action-free} datasets---such as human video data, which is often easier to collect than robot demonstrations---can provide additional supervision for the joint action-reasoning distribution $P_\theta$ which can in turn aid generalization. Specifically, we assume access to some action-free data $\tilde{\mathcal{D}}$ consisting of $M$ samples of $(\tilde{o_i}, \tilde{g_i}, \tilde{l_i}^1 \cdots \tilde{l_i}^{C_i})$. Here, sample $i$ includes the first $C_i \geq 1$ steps of language reasoning, where $C_i$ can vary between samples. For example, we might have varying levels of confidence in our full reasoning labeling pipeline for different subsets of our action-free data -- some samples might only be confident in the higher level reasoning steps (lower $C_i$) for example due to a large embodiment gap, while others might have high quality lower level reasoning (higher $C_i$). Importantly, this flexibility of reasoning labeling could enable our framework to incorporate vast scales of varying quality and embodiment reasoning data to improve \emph{each step} of the reasoning process independently from action prediction.

In this work, we optimize the objective above along with an auxiliary objective $\tilde{L}_{\texttt{reasoning}}(\theta)$ for the action-free data, defined similarly as follows:
\begin{align*}
    \tilde{L}_{\texttt{reasoning}}(\theta)& = \sum_i^M \sum_j^{C_i} \log P_\theta(\tilde{l_i}^j \mid \tilde{l_i}^1\dots \tilde{l_i}^{j-1},\tilde{o_i},\tilde{g_i})
\end{align*}

Note that since sample $i$ contains the first $C_i$ reasoning steps, we have enough information to model each of the $C_i$ reasoning steps conditioned on previous reasoning steps and the current observation and task. 

\subsection{Reasoning Steps in \ACRO}

While this setup can in principle work with different formulations of language reasoning steps, we instantiate our algorithm with the following reasoning steps from prior work \cite{zawalski2024robotic}:
\begin{itemize}
    \item \texttt{TaskPlan} ($l^1$): describes a list of subtasks to achieve $g$.
    \item \texttt{SubtaskReasoning} ($l^2$): reasons about which subtask currently needs to be executed in the plan.
    \item \texttt{Subtask} ($l^3$): predicts the subtask that currently needs to be executed.
    \item \texttt{MoveReasoning} ($l^4$): reasons about the motion needed to achieve the subtask in the scene.
    \item \texttt{MovePrimitive} ($l^5$): predicts a movement primitive in language.
    \item \texttt{GripperPosition} ($l^6$): predicts the pixel position of the end-effector.
    \item \texttt{VisibleObjects} ($l^7$): predicts the bounding box coordinates of objects in the scene.
    \item \texttt{Action} ($a$): predicts the low-level robot action as an end-effector position delta.
\end{itemize}

We note that these reasoning steps trace through information at an increasing amount of physical and spatial groundedness---beginning with high-level scene reasoning over tasks and subtasks, transitioning to reasoning over language motions, followed by spatial information about the gripper and objects, and concluding with the low-level robot action. We take advantage of this fact in designing a pipeline to label reasoning in action-free data, as we describe in the following section.


\subsection{Labeling Reasoning in Action-free Data}\label{sec:method:labeling}

In order to construct $\tilde{D}$---our dataset of observations, goals and action-free reasoning---we need to generate labels for the reasoning steps above from human videos. Our pipeline is similar to the automated procedure used by Embodied Chain-of-Thought (ECoT)~\cite{zawalski2024robotic} for generating reasoning over robot demonstrations, with some key modifications to handle human videos.
To obtain reasoning labels for robot demonstrations, ECoT first generates \texttt{GripperPositions} and \texttt{VisibleObjects} tags using off-the-shelf object detectors to obtain bounding boxes. Then, it extracts \texttt{MovePrimitive} (e.g. ``move to the left'') directly from actions using an automated heuristic. Conditioned on these more grounded reasoning steps $(l^5, l^6, l^7)$ and the image observation $o$, it queries Gemini~\cite{team2023gemini} to label the prior reasoning steps, from \texttt{TaskPlan} through \texttt{MoveReasoning} $(l^1, \dots, l^4)$. 

\begin{figure*}[!ht]
    \centering
    \includegraphics[width=\linewidth]{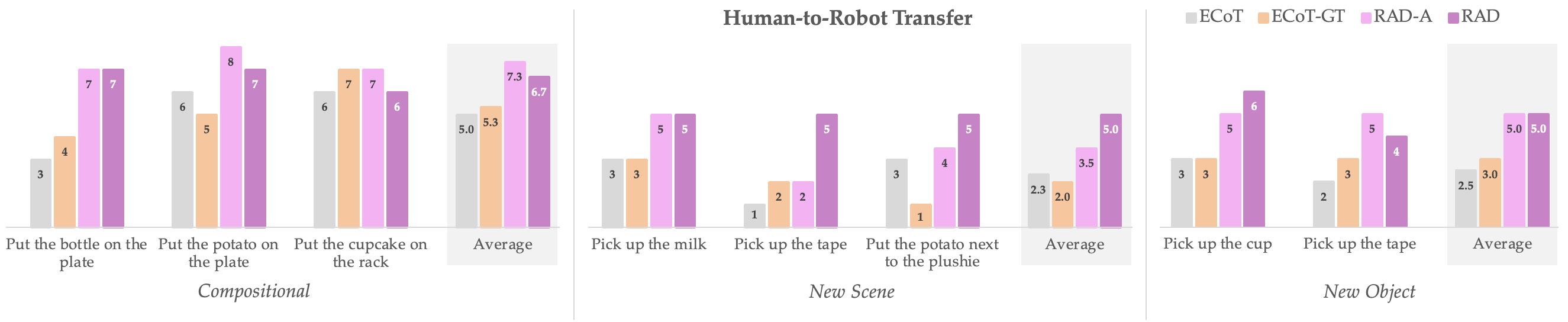}
    \caption{\ACRO outperforms baselines where human video data was trained on, but no new robot data was provided. \ACRO-A is \ACRO trained only on human video data for the given axis of generalization. ECoT-GT is finetuned on the same data as \ACRO, but only using human hand locations (and not the full reasoning data).}
    \label{fig:id_plot}
\end{figure*}

In the action-free setting with human videos, we note that we can still extract high-level reasoning with Gemini, as well as extract \texttt{VisibleObjects} with off-the-shelf object detectors. However, generating the more action-grounded reasoning steps is challenging: we can no longer extract \texttt{MovePrimitives} or \texttt{GripperPositions} automatically because we lack explicit action labels. In order to overcome this, we extract the \texttt{MovePrimitives} and \texttt{GripperPositions} using HaMeR~\cite{pavlakos2024reconstructing}, a hand keypoint and pose tracking method. Given these predictions, we can extract the \texttt{MovePrimitives} from changes in the hand pose information: first, we study each axis of the change in hand poses for each frame; then, we label the move primitive based on the dominant axis of motion. We find that this works reliably for tracking gripper and positional movement primitives, but is not as reliable for detecting rotational movement primitives.
We outline this labeling procedure in \cref{fig:labeling}.

\subsection{Training on Partial Reasoning Chains}\label{sec:method:training}

To train on mixtures of demonstration and action-free data, we use the ECoT and OpenVLA~\cite{zawalski2024robotic, kim2024openvla} architecture, which trains a pre-trained VLM transformer with 7B parameters to predict sequences of language reasoning and then action tokens. This model is pretrained on Internet-scale vision-language tasks, such as bounding box detection or object localization. Thus, it benefits from a strong vision and language priors. With ECoT and OpenVLA, it is then further trained on robot demonstration data, and in the case of ECoT, predicts language reasoning tokens prior to action tokens.
In \ACRO, we reuse this paradigm for the robot demonstration data, but for the new action-free data, our ``labels'' for training contain only reasoning as described in \cref{sec:method:labeling}.

%% file: sections/04_experiments.tex
In this section, we evaluate how \ACRO enables transfer from human videos to robot policies and generalization beyond settings in the human videos or robot demonstration data. Specifically, we seek to answer the following questions:

\smallskip \noindent\textbf{Q1} -- \textbf{Human-to-Robot Transfer:} Can \ACRO enable learning new tasks seen only in the human video data and not the robot demonstration data?


\noindent\textbf{Q2} -- \textbf{Reasoning Generalization:} Does reasoning in \ACRO enable generalization to novel tasks beyond both the robot demonstration data and human video data it was trained on?

\noindent\textbf{Q3} -- \textbf{Cross-Environment Transfer:} Can \ACRO learn new tasks from human video data in out-of-domain environments?


\subsection{Evaluating Generalization}\label{sec:experiments:evaluation}

Next, we discuss the environments, tasks, and model baselines we use to evaluate the reasoning generalization capabilities of \ACRO.

\smallskip \noindent \textbf{Real-World Environments}: We use a 6-DoF WidowX robot arm for our experiments. We perform all evaluations in \cref{sec:experiments:crossembodiment} and \cref{sec:experiments:generalization} on the Toy Sink setup from \cite{walke2023bridgedata}, to ensure fair comparison with existing pre-trained models. All human video data for \cref{sec:experiments:crossembodiment} and \cref{sec:experiments:generalization} was also collected in the Toy Sink setup (1616 demonstration videos), using both the standard Bridge V2 camera setup, as well as an additional camera for better hand tracking. Notably, the Bridge V2 setup is comprised of mostly miniature toy replicas of real world objects such as small kitchen supplies, blocks, and home supplies. Therefore, we also seek to assess how \ACRO responds to data from real-world human environments, and learns to interact with realistically sized objects. We thus collect data in two additional environments: a plain tabletop and a cluttered desk, as well as various real home and kitchen environments. This data was used to assess how \ACRO responds to data from unstructured environments in \cref{sec:experiments:crossenv}.

\smallskip \noindent \textbf{Generalization Tasks}: We evaluate \ACRO across a variety of generalization tasks. These tasks comprise three main axes of generalization:
\begin{enumerate}
    \item \textbf{Compositional Generalization}: In this axis, the objects, tasks, and scenes are all seen in pre-training data (Bridge V2 data), but not in those particular configurations. For example, pizza and salt both exist in Bridge V2, but salt is never placed on the pizza.
    \item \textbf{New Object Generalization}: This axis introduces unseen objects for known behaviors (e.g., \instr{pick cup} $\to$ \instr{pick plushie}).
    \item \textbf{New Scene Generalization}: This axis requires generalizing to novel backgrounds and distractor objects for seen tasks; for example, picking up a known object with a pot in the background.
\end{enumerate}

Note that the Compositional Generalization axis tests the model's ability to \emph{interpolate} the training data, while New Object and New Scene axes test the model's ability to \emph{extrapolate} from the training data. Exact tasks for each axis can be found in \cref{sec:appendix:results}.

\smallskip \noindent \textbf{Methods}: To test the efficacy of reasoning in learning from human video data, we evaluate the following models in our generalization scenarios.
\begin{enumerate}
    \item \textbf{Embodied Chain-of-Thought (ECoT)} \cite{zawalski2024robotic} A state-of-the-art action reasoning model trained on Bridge V2, but without any human video data.
    \item \textbf{ECoT w/ Gripper Tracking (ECoT-GT):} ECoT finetuned on the same human video data as \ACRO, but only generates the \texttt{GripperPosition} portion of the reasoning chain. This is analogous to how prior work learns from extracted pose information only in human videos, but does not extract higher level language reasoning \cite{papagiannis2024r+, lepertshadow, ren2025motion}.
    \item \textbf{\ACRO (Ours):} ECoT finetuned on the full chain of reasonings generated from human video data.
    \item \textbf{\ACRO-A (Ours):} Same as \ACRO, but trained on only human videos from one axis of generalization at a time (the axes are described in \cref{sec:experiments:evaluation}).
\end{enumerate}

\begin{figure*}[!ht]
    \centering
    \includegraphics[width=\linewidth]{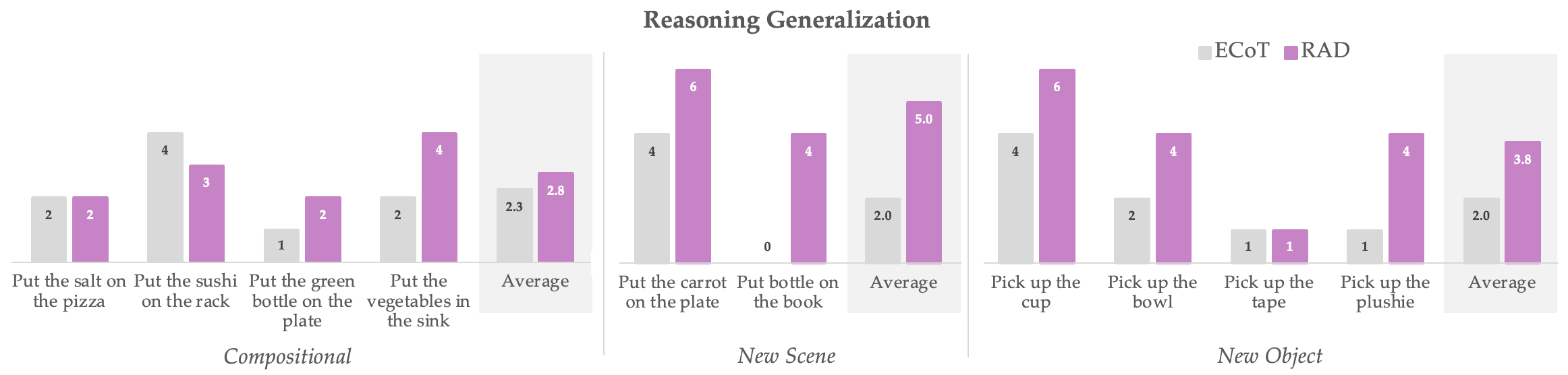}
    \caption{\ACRO compared to ECoT for tasks contained in neither human or robot data. \ACRO shows improved performance across all three axes of generalization.}
    \label{fig:ood_plot}
\end{figure*}

\subsection{Can \ACRO enable transfer from human-to-robot embodiments?}\label{sec:experiments:crossembodiment}

First, we assess if \ACRO can learn accurate reasonings and robot actions on new tasks that are present only in human video demonstrations. 
We train the axis-specific models (\ACRO-A) only on human video data for that axis (8-12 tasks with a total of 320-500 videos per axis). We evaluate these axis-specific models against zero-shot ECoT, as well as \ACRO (trained on human video data from all three axes) and ECoT-GT models trained on our full human video dataset.

In \cref{fig:id_plot}, we find that despite having no new robot demonstration data for these new tasks, \ACRO-A achieves consistently higher success rates than zero-shot ECoT and ECoT-GT across all areas of generalization (\textbf{Q1}). 

\smallskip \noindent \textbf{Compositional}: On compositionally new tasks, \ACRO-A outperforms ECoT by 23\% and ECoT-GT by 20\%. \ACRO outperforms ECoT and ECoT-GT by 17\% and 13\% respectively. Qualitatively, \ACRO models demonstrates significantly better reasoning capability, particularly in the second step of pick place tasks (such as  placing the object of interest in the desired location).

\smallskip \noindent \textbf{New Object}: On tasks with new objects, \ACRO and \ACRO-A both improves on ECoT and ECoT-GT by 25\% and 20\%, respectively. \ACRO models demonstrate substantially better ability to reason about grasp points on new objects, such as moving towards the sides of large cups instead of the middle. 

\smallskip \noindent \textbf{New Scene}: \ACRO models also substantially outperform baselines on novel scenes (containing  distractors and other scene modifications). \ACRO-A outperforms ECoT by 12\% and ECoT-GT by 15\%. The full \ACRO model had stronger performance, outperforming ECoT by 27\% and ECoT-GT by 30\% - potentially due to improves ability to ignore distractors from the larger dataset it was trained on. Reasoning traces on \ACRO models also appeared to be more accurate, with ECoT often becoming distracted and generating non-sensical reasonings.
These results indicate that augmenting chain-of-thought models with reasoning from human video data improves these models' ability to reason about and infer robot actions on previously unseen task configurations.

\subsection{Can \ACRO train more generalizable policies?}\label{sec:experiments:generalization}

Ultimately, training on large datasets of human video data should enable VLAs to generalize not only to human demonstrated tasks, but also to completely unseen scenarios. To explore if \ACRO enables training more general models, we evaluate our model against ECoT on 10 novel tasks (unseen in both human and robot data) comprising all three generalization axes. Results are presented in \cref{fig:ood_plot}.

\smallskip \noindent \textbf{Compositional}: On compositionally novel tasks, \ACRO outperforms ECoT by 5\%. \ACRO reasoned better than ECoT over multi-step tasks, such as knowing where to place the salt after picking it up.

\smallskip \noindent \textbf{New Object}: \ACRO substantially improves performance on tasks with unseen objects, such as bowls and large cups, despite not seeing such objects in human or robot training data. \ACRO achieves 30\% higher success compared to ECoT.

\smallskip \noindent \textbf{New Scene}: In novel scenes (environments with large distractors in the scene, such as cloth, pots, and a large plushie), \ACRO reached 18\% higher success rate than ECoT. Qualitatively, ECoT struggled to reason about the new scene and would often generate poor reasonings and execute seemingly random actions, whereas \ACRO generated correct reasoning which informed downstream action prediction.

\smallskip
This indicates that reasoning in \ACRO enables better generalization to a variety of unseen tasks, without training on any new human or robot data (\textbf{Q2}).

\begin{table}[ht]
\centering
\caption{Cross-Environment Transfer}
\label{tab:crossenv}
\begin{tabular}{llcc}
\toprule
\textbf{Task} & \textbf{Model} & \textbf{Success Rate} \\
\midrule
{\instr{pick up the cup}}
 & ECoT & 3/10 \\
 & \ACRO & 6/10 \\
 & ECoT-GT & 4/10 \\
\midrule
{\instr{put the sushi on the book}}
 & ECoT & 4.5/10 \\
 & \ACRO & 6.5/10 \\
 & ECoT-GT & 5/10 \\
 \midrule
{\instr{pick up the tiger}}
 & ECoT & 3/10 \\
 & \ACRO & 3/10 \\
 & ECoT-GT & 3/10 \\
 \midrule
{\instr{pick up the controller}}
 & ECoT & 2/10 \\
 & \ACRO & 3.5/10 \\
 & ECoT-GT & 2/10 \\
\bottomrule
\end{tabular}
\end{table}

\subsection{Can \ACRO leverage data from new environments?}\label{sec:experiments:crossenv}

To truly leverage large-scale video data, generalist robot policies must learn from demonstrations in diverse scenes. Thus, we first train \ACRO with human video data in unseen environments to see how well it can incorporate this data, and then we compare its performance to \ACRO trained on in-distribution human video data (i.e., same environment for both human video and robot evaluation).

\smallskip \noindent \textbf{Human Videos from New Environment}: We seek to understand how \ACRO responds to human video data collected outside the Bridge 
V2 environment. We first collect data for two unseen tasks in a new tabletop setup (unseen in Bridge V2 data). Then, we evaluate models trained on this new enviroment data in the original Bridge Toy Sink environment. In \cref{tab:crossenv}, we see that models trained on this data outperform ECoT by 16\% and ECoT-GT by 13\%. Similarly to \cref{sec:experiments:crossembodiment} and \cref{sec:experiments:generalization} \ACRO models showed significantly better ability to reason about grasp points, such as where to pick up the controller, despite the data being in a different environment (\textbf{Q3}).

\smallskip \noindent \textbf{In-distribution vs. Out-of-Distribution Human Data}: Next, we assess how \ACRO performance scales with increased data for the same tasks collected in-distribution (in the miniature Toy Sink setup) versus out-of-distribution (various real world kitchen and office environments). To do so, we collected 100 additional demos for the \instr{pick up the tape} task in the Toy Sink setup. We also collected 250 out-of-domain demos for \instr{pick up the tape} in novel environments such as real kitchens, countertops, and desks. Then, we trained \ACRO on two different data mixtures:
\begin{enumerate}
    \item The original \ACRO data mix (which already had 40 ``Pick up the tape'' demos) + in-distribution data and 
    \item The original \ACRO data mix + out-of-domain data.
\end{enumerate}

Results for both mixtures are shown in \cref{tab:tapescale}. We find that \ACRO models trained on both in-domain (+30\% success) and out-of-domain data (+25\% success) show improved performance over the original model (\textbf{Q3}). Qualitatively, \ACRO models were better able to reason about when to bring the gripper to the level of the tape, with ECoT models often moving to low and knocking over the tape, which is abnormally tall with respect to objects in Bridge V2.

\begin{table}[ht]
\centering
\caption{Data Scaling}
\label{tab:tapescale}
\begin{tabular}{llcc}
\toprule
\textbf{Data} & \textbf{Model} & \textbf{Success Rate} \\
\midrule
{\textbf{Original model (40 demos)}}
 & ECoT & 2/10 \\
 & ECoT-GT & 3/10 \\
 & \ACRO & 4/10 \\
 & \ACRO-A & 5/10 \\
\midrule
{\textbf{Same Environment (+100 ID demos)}}
 & \ACRO & 7/10 \\
 & ECoT-GT & 4/10 \\
\midrule
{\textbf{New Environments (+250 OOD demos)}}
 & \ACRO & 6.5/10 \\
 & ECoT-GT & 5/10 \\
\bottomrule
\end{tabular}
\end{table}

%% file: sections/05_discussion.tex
In this work we present \ACRO, a new way to train generalist robot policies from human video data. \ACRO learns to predict \textit{reasoning}, which can be labeled on both robot and human video data. We find that \ACRO enables VLAs to cross the embodiment gap, and to learn tasks represented in only human video data. Models trained with \ACRO are also able to generalize to completely unseen tasks (not present in either robot or human data). Finally, we find \ACRO responds positively to data from out-of-domain environments, enabling models to learn new tasks from environments completely separate from the target domain. These results demonstrate that \ACRO is a promising step towards training generalist robot policies, laying the groundwork for models that can leverage both robot data and large-scale human video data.

\smallskip \noindent \textbf{Limitations and Future Work}:
Our work demonstrates the promise of using human video data to improve generalization in robot policies; however, there are key challenges to address before scaling up the method to tap into larger and noisier datasets of human videos, such as those found on the Internet. For example, in the human video dataset we collected for this work, we limit the degrees-of-freedom that are expressed by the human hand to motions along Cartesian axes. As human hand pose estimation methods become more accurate, we anticipate that this limitation will be partially mitigated and allow us to leverage more natural videos of  human hands, as well as to expand the set of language motions in our labeling pipeline. Additionally, we scope our work to focus our study of generalization on pick-and-place tasks with rigid objects, characteristic of the tasks in prior work on reasoning-based imitation learning \cite{zawalski2024robotic}. Expanding the set of tasks to include more fine-grained and dexterous manipulation provides a rich area for future work.

%% file: sections/06_appendix.tex
We outline the dataset collection and reasoning generation procedure in \cref{sec:appendix:data}. The models, training procedure, and baselines are described in detail in \cref{sec:appendix:train}. Finally, \cref{sec:appendix:results} provides examples of results and description of reported success rates.

\subsection{Dataset Details}\label{sec:appendix:data}

\begin{figure}[!ht]
    \centering
    \includegraphics[width=\linewidth]{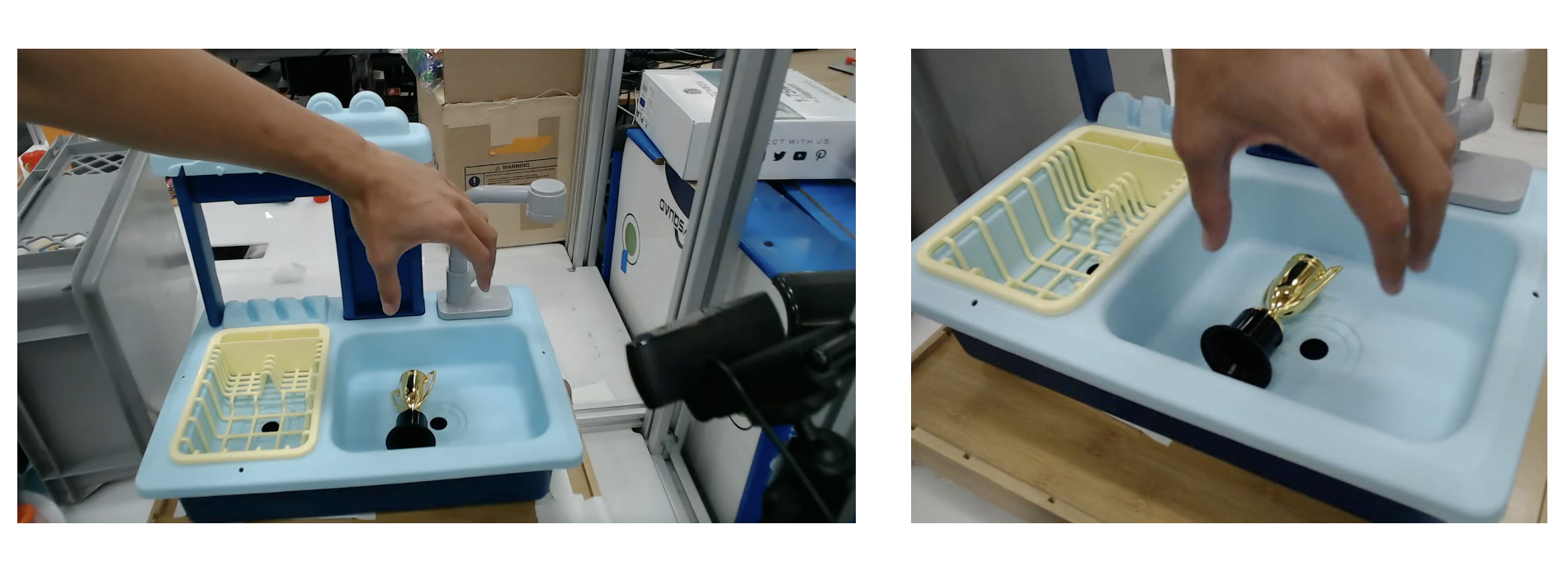}
    \caption{The main Bridge V2 perspective (right) versus the secondary perspective used for hand tracking (left).}
    \label{fig:povs}
\end{figure}

\begin{figure*}[!ht]
    \centering
    \includegraphics[width=\linewidth]{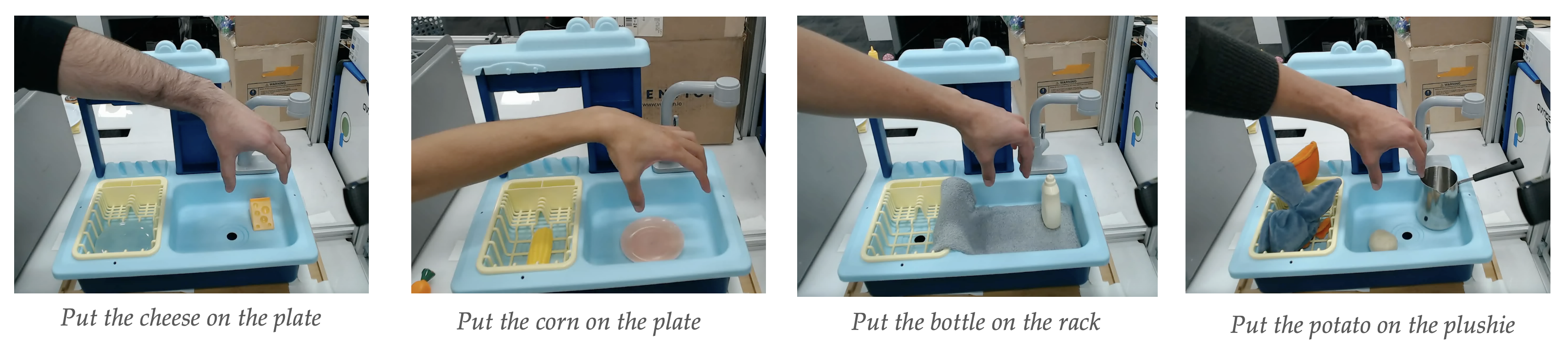}
    \caption{Example human video tasks collected.}
    \label{fig:data_col}
\end{figure*}

\smallskip \noindent \textbf{Data Collection}: Our main human video data collection was on the Bridge V2 Toy Sink setup. We aligned one camera based on the original Bridge V2 scene. We also set up a second camera from directly behind the WidowX gripper to better track hand movement as seen in \cref{fig:povs}. Example tasks are shown in \cref{fig:data_col}. We used HaMeR to track the hand using the secondary camera perspective. We used the average location of the thumb tip and index finger tip points tracked by HaMeR as the gripper location. Based on the delta gripper position between frames, we characterized every frame as ``stop", ``move forward", ``move backward", ``move left", ``move right", ``move up", or ``move down" movement primitives. We used the average distance between the thumb tip and index tip to determine ``close gripper" and ``open gripper" primitives. For reasoning generation on the human videos, we followed the the pipeline of \cite{zawalski2024robotic}, but used this HaMeR tracking in place of proprioception and SAM to generate movement primitives and gripper locations.

\begin{figure*}[!ht]
    \centering
    \includegraphics[width=\linewidth]{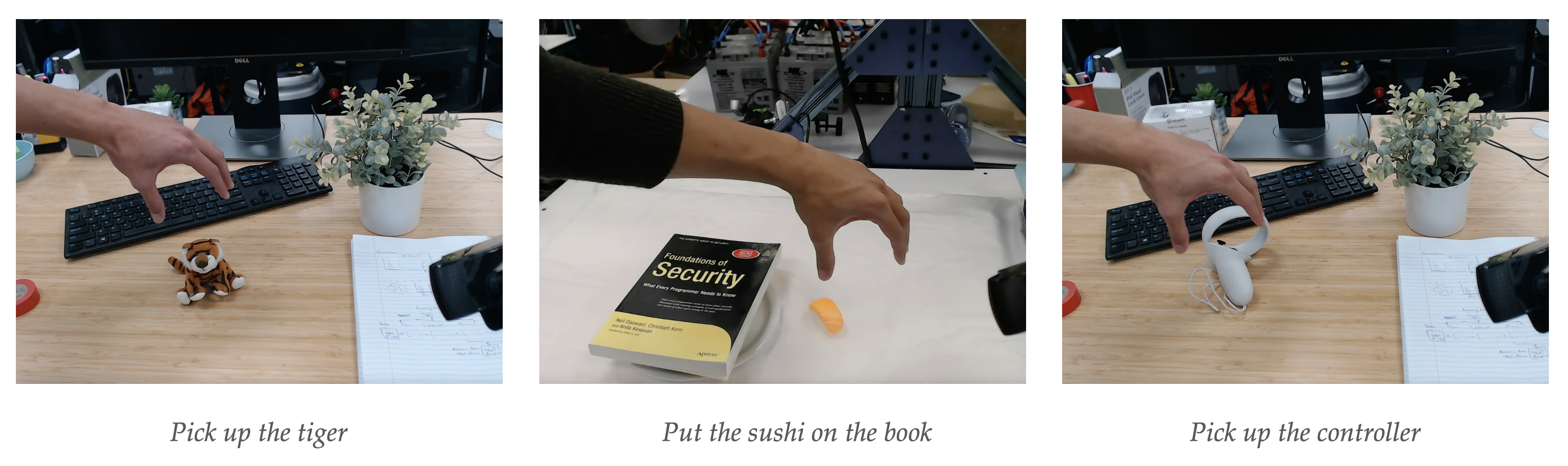}
    \caption{Task demonstrations collected in environments outside of Bridge V2 to assess how \ACRO responds to data from different types of scenes.}
    \label{fig:new_envs}
\end{figure*}

\begin{figure*}[!ht]
    \centering
    \includegraphics[width=\linewidth]{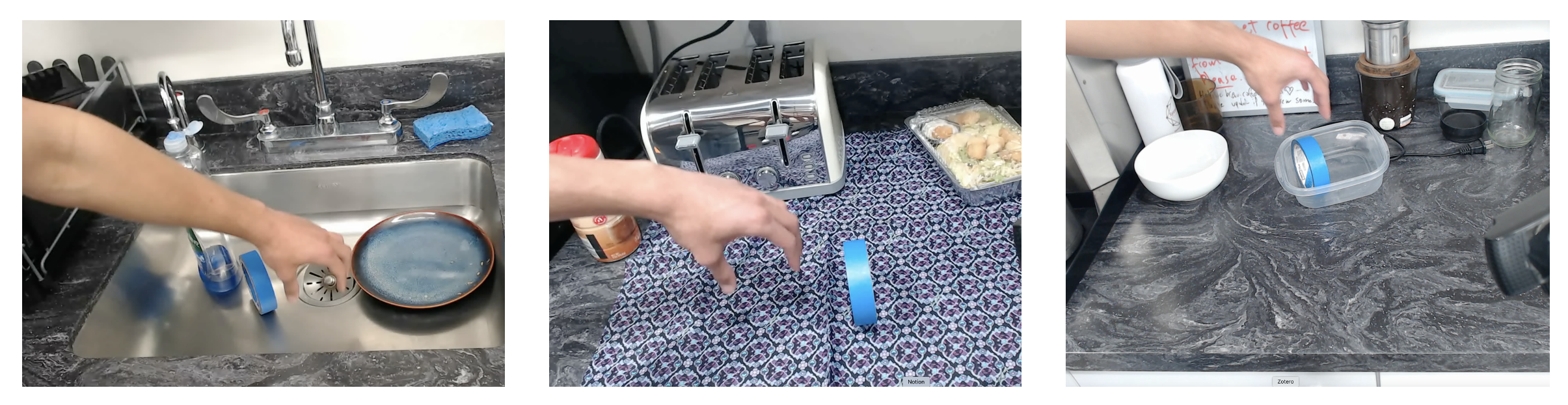}
    \caption{Real world environment data \ACRO is trained with for \cref{sec:experiments:crossenv}.}
    \label{fig:tape_envs}
\end{figure*}

\smallskip \noindent \textbf{Data Mixtures}: For \ACRO-A models in \cref{sec:experiments:crossembodiment} we collected 392 demonstrations for the compositional generalization dataset, 304 demonstrations for the new object dataset, and 280 demonstrations for the new scene dataset. The full \ACRO model as well as ECoT-GT model were both trained on all three of these datasets as well as 640 additional demos to make 1616 total demonstrations.

Data for \cref{tab:crossenv} was collected from two new tabletop environments as shown in \cref{fig:new_envs}. Each task in \cref{tab:crossenv} had 40 total demos collected. For \cref{tab:tapescale} we collected 100 additional demos in the Toy Sink setup for the ``in-distribution" evaluation. For the ``OOD" data, we collected 50 demos from 5 different scenes as show in \cref{fig:tape_envs}. 

\subsection{Training Details}\label{sec:appendix:train}

\ACRO uses the Prismatic VLM [35] architecture from OpenVLA \cite{kim2024openvla}, which fuses pre-trained SigLIP \cite{zhai2023sigmoid} and/or DinoV2 \cite{oquab2023dinov2} features for the visual encoder, and a LLaMA 2 7B \cite{touvron2023llama} language backbone. All models are fine-tuned to convergence with a learning rate of 2e-4, a LoRA batch size of 2, and anywhere from 2 to 8 GPUs (L40s or A40). Training of the ECoT-GT baseline is the same as \ACRO except the loss term for the stop token is omitted and we also adjust the query prompt from "What action should the robot take to [\textit{task}]?" to “Where is the robot hand in the image?”.

\begin{figure*}[!ht]
    \centering
    \includegraphics[width=\linewidth]{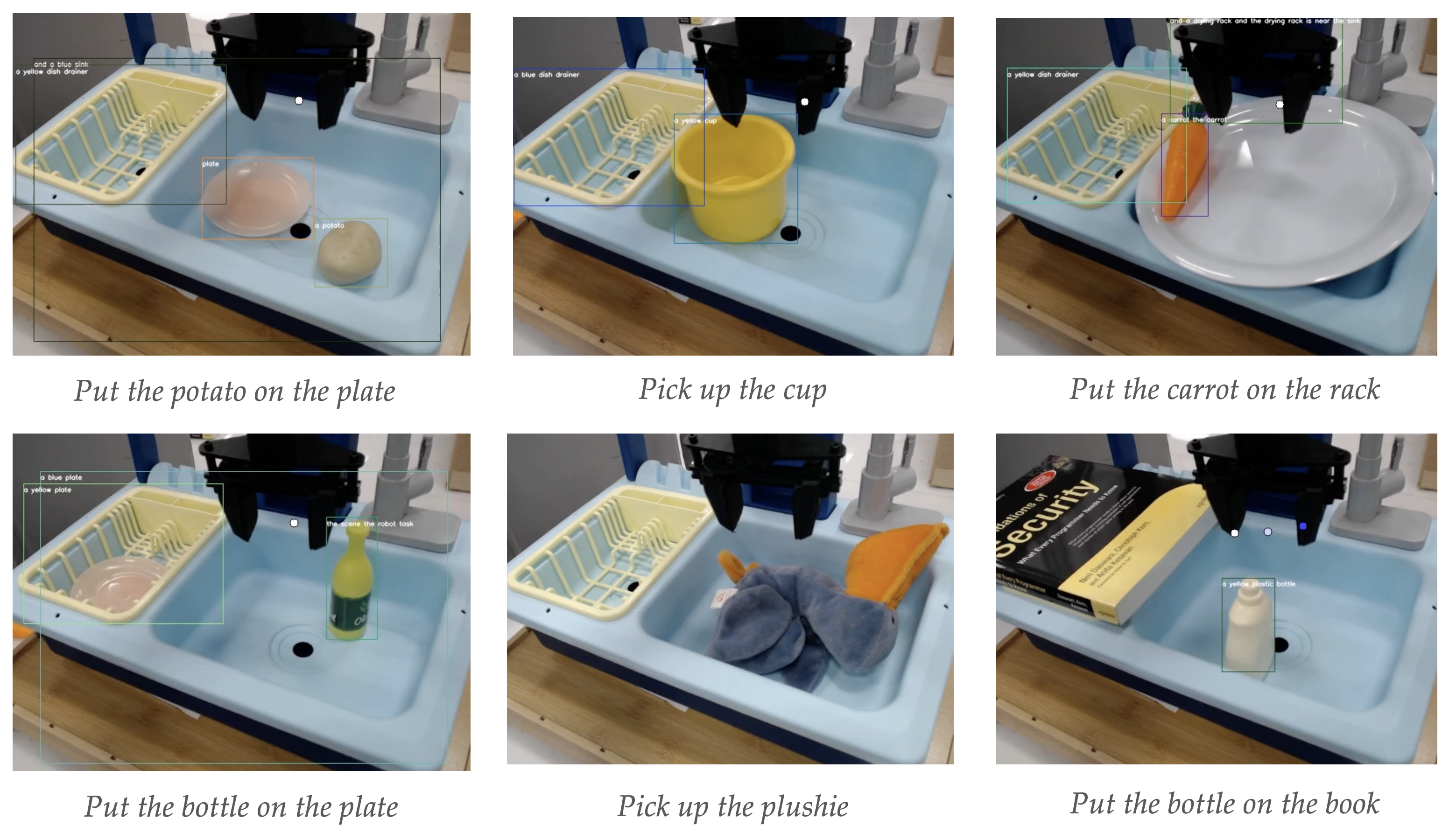}
    \caption{Example tasks for compositionally new tasks (left), new objects (middle), and new scenes (right).}
    \label{fig:tasks}
\end{figure*}

\subsection{Results}\label{sec:appendix:results}

Every task was evaluated 10 times. Objects were randomly placed throughout the scenes in a different spot for all 10 trials.
For pick and place tasks, partial credit (0.5) was given for successfully picking up the object, but placing in the wrong location. For pick objects, no partial credit was given except for the ``pick up the controller” task, which had an exceptionally high payload. Thus partial credit was given for grasping the object, even if the object slipped out of grasp upon being lifted.

%% file: main.bbl
\begin{thebibliography}{41}
\providecommand{\natexlab}[1]{#1}
\providecommand{\url}[1]{\texttt{#1}}
\expandafter\ifx\csname urlstyle\endcsname\relax
  \providecommand{\doi}[1]{doi: #1}\else
  \providecommand{\doi}{doi: \begingroup \urlstyle{rm}\Url}\fi

\bibitem[Bahl et~al.(2022)Bahl, Gupta, and Pathak]{bahl2022human}
Shikhar Bahl, Abhinav Gupta, and Deepak Pathak.
\newblock Human-to-robot imitation in the wild.
\newblock In \emph{Proceedings of Robotics: Science and Systems (RSS)}, 2022.

\bibitem[Belkhale et~al.(2024)Belkhale, Ding, Xiao, Sermanet, Vuong, Tompson, Chebotar, Dwibedi, and Sadigh]{belkhale2024rt}
Suneel Belkhale, Tianli Ding, Ted Xiao, Pierre Sermanet, Quon Vuong, Jonathan Tompson, Yevgen Chebotar, Debidatta Dwibedi, and Dorsa Sadigh.
\newblock {RT-H: Action hierarchies using language}.
\newblock In \emph{Proceedings of Robotics: Science and Systems (RSS)}, 2024.

\bibitem[Bharadhwaj et~al.(2024{\natexlab{a}})Bharadhwaj, Dwibedi, Gupta, Tulsiani, Doersch, Xiao, Shah, Xia, Sadigh, and Kirmani]{bharadhwaj2024gen2act}
Homanga Bharadhwaj, Debidatta Dwibedi, Abhinav Gupta, Shubham Tulsiani, Carl Doersch, Ted Xiao, Dhruv Shah, Fei Xia, Dorsa Sadigh, and Sean Kirmani.
\newblock Gen2act: Human video generation in novel scenarios enables generalizable robot manipulation.
\newblock \emph{arXiv preprint arXiv:2409.16283}, 2024{\natexlab{a}}.

\bibitem[Bharadhwaj et~al.(2024{\natexlab{b}})Bharadhwaj, Mottaghi, Gupta, and Tulsiani]{bharadhwaj2024track2act}
Homanga Bharadhwaj, Roozbeh Mottaghi, Abhinav Gupta, and Shubham Tulsiani.
\newblock Track2act: Predicting point tracks from internet videos enables diverse zero-shot robot manipulation.
\newblock \emph{arXiv preprint arXiv:2405.01527}, 2024{\natexlab{b}}.

\bibitem[Brohan et~al.(2022)Brohan, Brown, Carbajal, Chebotar, Dabis, Finn, Gopalakrishnan, Hausman, Herzog, Hsu, Ibarz, Ichter, Irpan, Jackson, Jesmonth, Joshi, Julian, Kalashnikov, Kuang, Leal, Lee, Levine, Lu, Malla, Manjunath, Mordatch, Nachum, Parada, Peralta, Perez, Pertsch, Quiambao, Rao, Ryoo, Salazar, Sanketi, Sayed, Singh, Sontakke, Stone, Tan, Tran, Vanhoucke, Vega, Vuong, Xia, Xiao, Xu, Xu, Yu, and Zitkovich]{rt12022}
Anthony Brohan, Noah Brown, Justice Carbajal, Yevgen Chebotar, Joseph Dabis, Chelsea Finn, Keerthana Gopalakrishnan, Karol Hausman, Alex Herzog, Jasmine Hsu, Julian Ibarz, Brian Ichter, Alex Irpan, Tomas Jackson, Sally Jesmonth, Nikhil Joshi, Ryan Julian, Dmitry Kalashnikov, Yuheng Kuang, Isabel Leal, Kuang-Huei Lee, Sergey Levine, Yao Lu, Utsav Malla, Deeksha Manjunath, Igor Mordatch, Ofir Nachum, Carolina Parada, Jodilyn Peralta, Emily Perez, Karl Pertsch, Jornell Quiambao, Kanishka Rao, Michael Ryoo, Grecia Salazar, Pannag Sanketi, Kevin Sayed, Jaspiar Singh, Sumedh Sontakke, Austin Stone, Clayton Tan, Huong Tran, Vincent Vanhoucke, Steve Vega, Quan Vuong, Fei Xia, Ted Xiao, Peng Xu, Sichun Xu, Tianhe Yu, and Brianna Zitkovich.
\newblock Rt-1: Robotics transformer for real-world control at scale.
\newblock \emph{arXiv}, 2022.

\bibitem[Brohan et~al.(2023{\natexlab{a}})Brohan, Brown, Carbajal, Chebotar, Chen, Choromanski, Ding, Driess, Dubey, Finn, Florence, Fu, Arenas, Gopalakrishnan, Han, Hausman, Herzog, Hsu, Ichter, Irpan, Joshi, Julian, Kalashnikov, Kuang, Leal, Lee, Lee, Levine, Lu, Michalewski, Mordatch, Pertsch, Rao, Reymann, Ryoo, Salazar, Sanketi, Sermanet, Singh, Singh, Soricut, Tran, Vanhoucke, Vuong, Wahid, Welker, Wohlhart, Wu, Xia, Xiao, Xu, Xu, Yu, and Zitkovich]{rt22023}
Anthony Brohan, Noah Brown, Justice Carbajal, Yevgen Chebotar, Xi~Chen, Krzysztof Choromanski, Tianli Ding, Danny Driess, Avinava Dubey, Chelsea Finn, Pete Florence, Chuyuan Fu, Montse~Gonzalez Arenas, Keerthana Gopalakrishnan, Kehang Han, Karol Hausman, Alex Herzog, Jasmine Hsu, Brian Ichter, Alex Irpan, Nikhil Joshi, Ryan Julian, Dmitry Kalashnikov, Yuheng Kuang, Isabel Leal, Lisa Lee, Tsang-Wei~Edward Lee, Sergey Levine, Yao Lu, Henryk Michalewski, Igor Mordatch, Karl Pertsch, Kanishka Rao, Krista Reymann, Michael Ryoo, Grecia Salazar, Pannag Sanketi, Pierre Sermanet, Jaspiar Singh, Anikait Singh, Radu Soricut, Huong Tran, Vincent Vanhoucke, Quan Vuong, Ayzaan Wahid, Stefan Welker, Paul Wohlhart, Jialin Wu, Fei Xia, Ted Xiao, Peng Xu, Sichun Xu, Tianhe Yu, and Brianna Zitkovich.
\newblock {RT-2: Vision-Language-Action Models Transfer Web Knowledge to Robotic Control}.
\newblock In \emph{arXiv}, 2023{\natexlab{a}}.

\bibitem[Brohan et~al.(2023{\natexlab{b}})Brohan, Chebotar, Finn, Hausman, Herzog, Ho, Ibarz, Irpan, Jang, Julian, et~al.]{brohan2023do}
Anthony Brohan, Yevgen Chebotar, Chelsea Finn, Karol Hausman, Alexander Herzog, Daniel Ho, Julian Ibarz, Alex Irpan, Eric Jang, Ryan Julian, et~al.
\newblock {Do as I can, not as I say: Grounding language in robotic affordances}.
\newblock In \emph{Conference on Robot Learning}, pages 287--318, 2023{\natexlab{b}}.

\bibitem[Chen et~al.(2021)Chen, Nair, and Finn]{chen2021learning}
Annie~S Chen, Suraj Nair, and Chelsea Finn.
\newblock Learning generalizable robotic reward functions from" in-the-wild" human videos.
\newblock In \emph{Proceedings of Robotics: Science and Systems (RSS)}, 2021.

\bibitem[Chen et~al.(2024)Chen, Djolonga, Padlewski, Mustafa, Changpinyo, Wu, Riquelme, Goodman, Wang, Tay, Shakeri, Dehghani, Salz, Lučić, Tschannen, Nagrani, Hu, Joshi, Pang, Montgomery, Pietrzyk, Ritter, Piergiovanni, Minderer, Pavetić, Waters, Li, Alabdulmohsin, Beyer, Amelot, Lee, Steiner, Li, Keysers, Arnab, Xu, Rong, Kolesnikov, Seyedhosseini, Angelova, Zhai, Houlsby, and Soricut]{chen2024palix}
Xi~Chen, Josip Djolonga, Piotr Padlewski, Basil Mustafa, Beer Changpinyo, Jialin Wu, Carlos Riquelme, Sebastian Goodman, Xiao Wang, Yi~Tay, Siamak Shakeri, Mostafa Dehghani, Daniel Salz, Mario Lučić, Michael Tschannen, Arsha Nagrani, Hexiang~(Frank) Hu, Mandar Joshi, Bo~Pang, Ceslee Montgomery, Paulina Pietrzyk, Marvin Ritter, AJ~Piergiovanni, Matthias Minderer, Filip Pavetić, Austin Waters, Gang Li, Ibrahim Alabdulmohsin, Lucas Beyer, Julien Amelot, Kenton Lee, Andreas Steiner, Yang Li, Daniel Keysers, Anurag Arnab, Yuanzhong Xu, Keran Rong, Alexander Kolesnikov, Mojtaba Seyedhosseini, Anelia Angelova, Xiaohua Zhai, Neil Houlsby, and Radu Soricut.
\newblock Pali-x: On scaling up a multilingual vision and language model.
\newblock In \emph{Conference on Computer Vision and Pattern Recognition (CVPR)}, 2024.

\bibitem[Collaboration et~al.(2024)Collaboration, O'Neill, Rehman, Gupta, Maddukuri, Gupta, Padalkar, Lee, Pooley, Gupta, Mandlekar, Jain, Tung, Bewley, Herzog, Irpan, Khazatsky, Rai, Gupta, Wang, Kolobov, Singh, Garg, Kembhavi, Xie, Brohan, Raffin, Sharma, Yavary, Jain, Balakrishna, Wahid, Burgess-Limerick, Kim, Schölkopf, Wulfe, Ichter, Lu, Xu, Le, Finn, Wang, Xu, Chi, Huang, Chan, Agia, Pan, Fu, Devin, Xu, Morton, Driess, Chen, Pathak, Shah, Büchler, Jayaraman, Kalashnikov, Sadigh, Johns, Foster, Liu, Ceola, Xia, Zhao, Frujeri, Stulp, Zhou, Sukhatme, Salhotra, Yan, Feng, Schiavi, Berseth, Kahn, Yang, Wang, Su, Fang, Shi, Bao, Amor, Christensen, Furuta, Bharadhwaj, Walke, Fang, Ha, Mordatch, Radosavovic, Leal, Liang, Abou-Chakra, Kim, Drake, Peters, Schneider, Hsu, Vakil, Bohg, Bingham, Wu, Gao, Hu, Wu, Wu, Sun, Luo, Gu, Tan, Oh, Wu, Lu, Yang, Malik, Silvério, Hejna, Booher, Tompson, Yang, Salvador, Lim, Han, Wang, Rao, Pertsch, Hausman, Go, Gopalakrishnan, Goldberg, Byrne, Oslund, Kawaharazuka, Black,
  Lin, Zhang, Ehsani, Lekkala, Ellis, Rana, Srinivasan, Fang, Singh, Zeng, Hatch, Hsu, Itti, Chen, Pinto, Fei-Fei, Tan, Fan, Ott, Lee, Weihs, Chen, Lepert, Memmel, Tomizuka, Itkina, Castro, Spero, Du, Ahn, Yip, Zhang, Ding, Heo, Srirama, Sharma, Kim, Kanazawa, Hansen, Heess, Joshi, Suenderhauf, Liu, Palo, Shafiullah, Mees, Kroemer, Bastani, Sanketi, Miller, Yin, Wohlhart, Xu, Fagan, Mitrano, Sermanet, Abbeel, Sundaresan, Chen, Vuong, Rafailov, Tian, Doshi, Mart{'i}n-Mart{'i}n, Baijal, Scalise, Hendrix, Lin, Qian, Zhang, Mendonca, Shah, Hoque, Julian, Bustamante, Kirmani, Levine, Lin, Moore, Bahl, Dass, Sonawani, Tulsiani, Song, Xu, Haldar, Karamcheti, Adebola, Guist, Nasiriany, Schaal, Welker, Tian, Ramamoorthy, Dasari, Belkhale, Park, Nair, Mirchandani, Osa, Gupta, Harada, Matsushima, Xiao, Kollar, Yu, Ding, Davchev, Zhao, Armstrong, Darrell, Chung, Jain, Kumar, Vanhoucke, Zhan, Zhou, Burgard, Chen, Chen, Wang, Zhu, Geng, Liu, Liangwei, Li, Pang, Lu, Ma, Kim, Chebotar, Zhou, Zhu, Wu, Xu, Wang, Bisk, Dou,
  Cho, Lee, Cui, Cao, Wu, Tang, Zhu, Zhang, Jiang, Li, Li, Iwasawa, Matsuo, Ma, Xu, Cui, Zhang, Fu, and Lin]{oxe2024}
Open X-Embodiment Collaboration, Abby O'Neill, Abdul Rehman, Abhinav Gupta, Abhiram Maddukuri, Abhishek Gupta, Abhishek Padalkar, Abraham Lee, Acorn Pooley, Agrim Gupta, Ajay Mandlekar, Ajinkya Jain, Albert Tung, Alex Bewley, Alex Herzog, Alex Irpan, Alexander Khazatsky, Anant Rai, Anchit Gupta, Andrew Wang, Andrey Kolobov, Anikait Singh, Animesh Garg, Aniruddha Kembhavi, Annie Xie, Anthony Brohan, Antonin Raffin, Archit Sharma, Arefeh Yavary, Arhan Jain, Ashwin Balakrishna, Ayzaan Wahid, Ben Burgess-Limerick, Beomjoon Kim, Bernhard Schölkopf, Blake Wulfe, Brian Ichter, Cewu Lu, Charles Xu, Charlotte Le, Chelsea Finn, Chen Wang, Chenfeng Xu, Cheng Chi, Chenguang Huang, Christine Chan, Christopher Agia, Chuer Pan, Chuyuan Fu, Coline Devin, Danfei Xu, Daniel Morton, Danny Driess, Daphne Chen, Deepak Pathak, Dhruv Shah, Dieter Büchler, Dinesh Jayaraman, Dmitry Kalashnikov, Dorsa Sadigh, Edward Johns, Ethan Foster, Fangchen Liu, Federico Ceola, Fei Xia, Feiyu Zhao, Felipe~Vieira Frujeri, Freek Stulp, Gaoyue
  Zhou, Gaurav~S. Sukhatme, Gautam Salhotra, Ge~Yan, Gilbert Feng, Giulio Schiavi, Glen Berseth, Gregory Kahn, Guangwen Yang, Guanzhi Wang, Hao Su, Hao-Shu Fang, Haochen Shi, Henghui Bao, Heni~Ben Amor, Henrik~I Christensen, Hiroki Furuta, Homanga Bharadhwaj, Homer Walke, Hongjie Fang, Huy Ha, Igor Mordatch, Ilija Radosavovic, Isabel Leal, Jacky Liang, Jad Abou-Chakra, Jaehyung Kim, Jaimyn Drake, Jan Peters, Jan Schneider, Jasmine Hsu, Jay Vakil, Jeannette Bohg, Jeffrey Bingham, Jeffrey Wu, Jensen Gao, Jiaheng Hu, Jiajun Wu, Jialin Wu, Jiankai Sun, Jianlan Luo, Jiayuan Gu, Jie Tan, Jihoon Oh, Jimmy Wu, Jingpei Lu, Jingyun Yang, Jitendra Malik, João Silvério, Joey Hejna, Jonathan Booher, Jonathan Tompson, Jonathan Yang, Jordi Salvador, Joseph~J. Lim, Junhyek Han, Kaiyuan Wang, Kanishka Rao, Karl Pertsch, Karol Hausman, Keegan Go, Keerthana Gopalakrishnan, Ken Goldberg, Kendra Byrne, Kenneth Oslund, Kento Kawaharazuka, Kevin Black, Kevin Lin, Kevin Zhang, Kiana Ehsani, Kiran Lekkala, Kirsty Ellis, Krishan
  Rana, Krishnan Srinivasan, Kuan Fang, Kunal~Pratap Singh, Kuo-Hao Zeng, Kyle Hatch, Kyle Hsu, Laurent Itti, Lawrence~Yunliang Chen, Lerrel Pinto, Li~Fei-Fei, Liam Tan, Linxi~"Jim" Fan, Lionel Ott, Lisa Lee, Luca Weihs, Magnum Chen, Marion Lepert, Marius Memmel, Masayoshi Tomizuka, Masha Itkina, Mateo~Guaman Castro, Max Spero, Maximilian Du, Michael Ahn, Michael~C. Yip, Mingtong Zhang, Mingyu Ding, Minho Heo, Mohan~Kumar Srirama, Mohit Sharma, Moo~Jin Kim, Naoaki Kanazawa, Nicklas Hansen, Nicolas Heess, Nikhil~J Joshi, Niko Suenderhauf, Ning Liu, Norman~Di Palo, Nur Muhammad~Mahi Shafiullah, Oier Mees, Oliver Kroemer, Osbert Bastani, Pannag~R Sanketi, Patrick~"Tree" Miller, Patrick Yin, Paul Wohlhart, Peng Xu, Peter~David Fagan, Peter Mitrano, Pierre Sermanet, Pieter Abbeel, Priya Sundaresan, Qiuyu Chen, Quan Vuong, Rafael Rafailov, Ran Tian, Ria Doshi, Roberto Mart{'i}n-Mart{'i}n, Rohan Baijal, Rosario Scalise, Rose Hendrix, Roy Lin, Runjia Qian, Ruohan Zhang, Russell Mendonca, Rutav Shah, Ryan Hoque, Ryan
  Julian, Samuel Bustamante, Sean Kirmani, Sergey Levine, Shan Lin, Sherry Moore, Shikhar Bahl, Shivin Dass, Shubham Sonawani, Shubham Tulsiani, Shuran Song, Sichun Xu, Siddhant Haldar, Siddharth Karamcheti, Simeon Adebola, Simon Guist, Soroush Nasiriany, Stefan Schaal, Stefan Welker, Stephen Tian, Subramanian Ramamoorthy, Sudeep Dasari, Suneel Belkhale, Sungjae Park, Suraj Nair, Suvir Mirchandani, Takayuki Osa, Tanmay Gupta, Tatsuya Harada, Tatsuya Matsushima, Ted Xiao, Thomas Kollar, Tianhe Yu, Tianli Ding, Todor Davchev, Tony~Z. Zhao, Travis Armstrong, Trevor Darrell, Trinity Chung, Vidhi Jain, Vikash Kumar, Vincent Vanhoucke, Wei Zhan, Wenxuan Zhou, Wolfram Burgard, Xi~Chen, Xiangyu Chen, Xiaolong Wang, Xinghao Zhu, Xinyang Geng, Xiyuan Liu, Xu~Liangwei, Xuanlin Li, Yansong Pang, Yao Lu, Yecheng~Jason Ma, Yejin Kim, Yevgen Chebotar, Yifan Zhou, Yifeng Zhu, Yilin Wu, Ying Xu, Yixuan Wang, Yonatan Bisk, Yongqiang Dou, Yoonyoung Cho, Youngwoon Lee, Yuchen Cui, Yue Cao, Yueh-Hua Wu, Yujin Tang, Yuke Zhu,
  Yunchu Zhang, Yunfan Jiang, Yunshuang Li, Yunzhu Li, Yusuke Iwasawa, Yutaka Matsuo, Zehan Ma, Zhuo Xu, Zichen~Jeff Cui, Zichen Zhang, Zipeng Fu, and Zipeng Lin.
\newblock Open {X-E}mbodiment: Robotic learning datasets and {RT-X} models.
\newblock In \emph{International Conference on Robotics and Automation (ICRA)}, 2024.

\bibitem[Huang et~al.(2023)Huang, Xia, Xiao, Chan, Liang, Florence, Zeng, Tompson, Mordatch, Chebotar, et~al.]{huang2023inner}
Wenlong Huang, Fei Xia, Ted Xiao, Harris Chan, Jacky Liang, Pete Florence, Andy Zeng, Jonathan Tompson, Igor Mordatch, Yevgen Chebotar, et~al.
\newblock Inner monologue: Embodied reasoning through planning with language models.
\newblock In \emph{Conference on Robot Learning}, 2023.

\bibitem[Jain et~al.(2024)Jain, Attarian, Joshi, Wahid, Driess, Vuong, Sanketi, Sermanet, Welker, Chan, et~al.]{jain2024vid2robot}
Vidhi Jain, Maria Attarian, Nikhil~J Joshi, Ayzaan Wahid, Danny Driess, Quan Vuong, Pannag~R Sanketi, Pierre Sermanet, Stefan Welker, Christine Chan, et~al.
\newblock Vid2robot: End-to-end video-conditioned policy learning with cross-attention transformers.
\newblock \emph{arXiv preprint arXiv:2403.12943}, 2024.

\bibitem[Jang et~al.(2022)Jang, Irpan, Khansari, Kappler, Ebert, Lynch, Levine, and Finn]{jang2022bc}
Eric Jang, Alex Irpan, Mohi Khansari, Daniel Kappler, Frederik Ebert, Corey Lynch, Sergey Levine, and Chelsea Finn.
\newblock {BC-Z: Zero-shot task generalization with robotic imitation learning}.
\newblock In \emph{Conference on Robot Learning (CoRL)}, 2022.

\bibitem[Karamcheti et~al.(2023)Karamcheti, Nair, Chen, Kollar, Finn, Sadigh, and Liang]{karamcheti2023language}
Siddharth Karamcheti, Suraj Nair, Annie~S Chen, Thomas Kollar, Chelsea Finn, Dorsa Sadigh, and Percy Liang.
\newblock Language-driven representation learning for robotics.
\newblock \emph{arXiv preprint arXiv:2302.12766}, 2023.

\bibitem[Khazatsky et~al.(2024)Khazatsky, Pertsch, Nair, Balakrishna, Dasari, Karamcheti, Nasiriany, Srirama, Chen, Ellis, et~al.]{khazatsky2024droid}
Alexander Khazatsky, Karl Pertsch, Suraj Nair, Ashwin Balakrishna, Sudeep Dasari, Siddharth Karamcheti, Soroush Nasiriany, Mohan~Kumar Srirama, Lawrence~Yunliang Chen, Kirsty Ellis, et~al.
\newblock Droid: A large-scale in-the-wild robot manipulation dataset.
\newblock \emph{arXiv preprint arXiv:2403.12945}, 2024.

\bibitem[Kim et~al.(2024)Kim, Pertsch, Karamcheti, Xiao, Balakrishna, Nair, Rafailov, Foster, Lam, Sanketi, et~al.]{kim2024openvla}
Moo~Jin Kim, Karl Pertsch, Siddharth Karamcheti, Ted Xiao, Ashwin Balakrishna, Suraj Nair, Rafael Rafailov, Ethan Foster, Grace Lam, Pannag Sanketi, et~al.
\newblock Openvla: An open-source vision-language-action model.
\newblock \emph{Conference on Robot Learning (CoRL)}, 2024.

\bibitem[Lepert et~al.()Lepert, Doshi, and Bohg]{lepertshadow}
Marion Lepert, Ria Doshi, and Jeannette Bohg.
\newblock Shadow: Leveraging segmentation masks for cross-embodiment policy transfer.
\newblock In \emph{8th Annual Conference on Robot Learning}.

\bibitem[Mandikal and Grauman(2022)]{mandikal2022dexvip}
Priyanka Mandikal and Kristen Grauman.
\newblock {DexVIP: Learning dexterous grasping with human hand pose priors from video}.
\newblock In \emph{Conference on Robot Learning (CoRL)}, 2022.

\bibitem[Nair et~al.(2022)Nair, Rajeswaran, Kumar, Finn, and Gupta]{nair2022r3m}
Suraj Nair, Aravind Rajeswaran, Vikash Kumar, Chelsea Finn, and Abhinav Gupta.
\newblock R3m: A universal visual representation for robot manipulation.
\newblock In \emph{Conference on Robot Learning (CoRL)}, 2022.

\bibitem[Oquab et~al.(2023)Oquab, Darcet, Moutakanni, Vo, Szafraniec, Khalidov, Fernandez, Haziza, Massa, El-Nouby, et~al.]{oquab2023dinov2}
Maxime Oquab, Timoth{\'e}e Darcet, Th{\'e}o Moutakanni, Huy Vo, Marc Szafraniec, Vasil Khalidov, Pierre Fernandez, Daniel Haziza, Francisco Massa, Alaaeldin El-Nouby, et~al.
\newblock Dinov2: Learning robust visual features without supervision.
\newblock \emph{arXiv preprint arXiv:2304.07193}, 2023.

\bibitem[Papagiannis et~al.(2024)Papagiannis, Di~Palo, Vitiello, and Johns]{papagiannis2024r+}
Georgios Papagiannis, Norman Di~Palo, Pietro Vitiello, and Edward Johns.
\newblock R+ x: Retrieval and execution from everyday human videos.
\newblock \emph{arXiv preprint arXiv:2407.12957}, 2024.

\bibitem[Pavlakos et~al.(2024)Pavlakos, Shan, Radosavovic, Kanazawa, Fouhey, and Malik]{pavlakos2024reconstructing}
Georgios Pavlakos, Dandan Shan, Ilija Radosavovic, Angjoo Kanazawa, David Fouhey, and Jitendra Malik.
\newblock Reconstructing hands in 3{D} with transformers.
\newblock In \emph{Conference on Computer Vision and Pattern Recognition (CVPR)}, 2024.

\bibitem[Qin et~al.(2022)Qin, Wu, Liu, Jiang, Yang, Fu, and Wang]{qin2022dexmv}
Yuzhe Qin, Yueh-Hua Wu, Shaowei Liu, Hanwen Jiang, Ruihan Yang, Yang Fu, and Xiaolong Wang.
\newblock {DexMV: Imitation learning for dexterous manipulation from human videos}.
\newblock In \emph{European Conference on Computer Vision}, 2022.

\bibitem[Ren et~al.(2025)Ren, Sundaresan, Sadigh, Choudhury, and Bohg]{ren2025motion}
Juntao Ren, Priya Sundaresan, Dorsa Sadigh, Sanjiban Choudhury, and Jeannette Bohg.
\newblock Motion tracks: A unified representation for human-robot transfer in few-shot imitation learning.
\newblock \emph{arXiv preprint arXiv:2501.06994}, 2025.

\bibitem[Shao et~al.(2021)Shao, Migimatsu, Zhang, Yang, and Bohg]{shao2021concept2robot}
Lin Shao, Toki Migimatsu, Qiang Zhang, Karen Yang, and Jeannette Bohg.
\newblock {Concept2Robot: Learning manipulation concepts from instructions and human demonstrations}.
\newblock \emph{The International Journal of Robotics Research (IJRR)}, 2021.

\bibitem[Sharma et~al.(2019)Sharma, Pathak, and Gupta]{sharma2019third}
Pratyusha Sharma, Deepak Pathak, and Abhinav Gupta.
\newblock Third-person visual imitation learning via decoupled hierarchical controller.
\newblock In \emph{Advances in Neural Information Processing Systems (NeurIPS)}, 2019.

\bibitem[Shaw et~al.(2023)Shaw, Bahl, and Pathak]{shaw2023videodex}
Kenneth Shaw, Shikhar Bahl, and Deepak Pathak.
\newblock Videodex: Learning dexterity from internet videos.
\newblock In \emph{Conference on Robot Learning (CoRL)}, 2023.

\bibitem[Shi et~al.(2024)Shi, Hu, Zhao, Sharma, Pertsch, Luo, Levine, and Finn]{shi2024yell}
Lucy~Xiaoyang Shi, Zheyuan Hu, Tony~Z Zhao, Archit Sharma, Karl Pertsch, Jianlan Luo, Sergey Levine, and Chelsea Finn.
\newblock {Yell at your robot: Improving on-the-fly from language corrections}.
\newblock In \emph{Proceedings of Robotics: Science and Systems (RSS)}, 2024.

\bibitem[Smith et~al.(2019)Smith, Dhawan, Zhang, Abbeel, and Levine]{smith2019avid}
Laura Smith, Nikita Dhawan, Marvin Zhang, Pieter Abbeel, and Sergey Levine.
\newblock Avid: Learning multi-stage tasks via pixel-level translation of human videos.
\newblock In \emph{Proceedings of Robotics: Science and Systems (RSS)}, 2019.

\bibitem[Stepputtis et~al.(2020)Stepputtis, Campbell, Phielipp, Lee, Baral, and Ben~Amor]{stepputtis2020language}
Simon Stepputtis, Joseph Campbell, Mariano Phielipp, Stefan Lee, Chitta Baral, and Heni Ben~Amor.
\newblock Language-conditioned imitation learning for robot manipulation tasks.
\newblock \emph{Advances in Neural Information Processing Systems (NeurIPS)}, 2020.

\bibitem[Team et~al.(2023)Team, Anil, Borgeaud, Alayrac, Yu, Soricut, Schalkwyk, Dai, Hauth, Millican, et~al.]{team2023gemini}
Gemini Team, Rohan Anil, Sebastian Borgeaud, Jean-Baptiste Alayrac, Jiahui Yu, Radu Soricut, Johan Schalkwyk, Andrew~M Dai, Anja Hauth, Katie Millican, et~al.
\newblock Gemini: a family of highly capable multimodal models.
\newblock \emph{arXiv}, 2023.

\bibitem[Team et~al.(2024)Team, Ghosh, Walke, Pertsch, Black, Mees, Dasari, Hejna, Kreiman, Xu, et~al.]{team2024octo}
Octo~Model Team, Dibya Ghosh, Homer Walke, Karl Pertsch, Kevin Black, Oier Mees, Sudeep Dasari, Joey Hejna, Tobias Kreiman, Charles Xu, et~al.
\newblock Octo: An open-source generalist robot policy.
\newblock \emph{arXiv preprint arXiv:2405.12213}, 2024.

\bibitem[Touvron et~al.(2023)Touvron, Martin, Stone, Albert, Almahairi, Babaei, Bashlykov, Batra, Bhargava, Bhosale, et~al.]{touvron2023llama}
Hugo Touvron, Louis Martin, Kevin Stone, Peter Albert, Amjad Almahairi, Yasmine Babaei, Nikolay Bashlykov, Soumya Batra, Prajjwal Bhargava, Shruti Bhosale, et~al.
\newblock Llama 2: Open foundation and fine-tuned chat models.
\newblock \emph{arXiv preprint arXiv:2307.09288}, 2023.

\bibitem[Walke et~al.(2023)Walke, Black, Zhao, Vuong, Zheng, Hansen-Estruch, He, Myers, Kim, Du, et~al.]{walke2023bridgedata}
Homer~Rich Walke, Kevin Black, Tony~Z Zhao, Quan Vuong, Chongyi Zheng, Philippe Hansen-Estruch, Andre~Wang He, Vivek Myers, Moo~Jin Kim, Max Du, et~al.
\newblock Bridgedata v2: A dataset for robot learning at scale.
\newblock In \emph{Conference on Robot Learning}, pages 1723--1736. PMLR, 2023.

\bibitem[Wang et~al.(2023)Wang, Fan, Sun, Zhang, Fei-Fei, Xu, Zhu, and Anandkumar]{wang2023mimicplay}
Chen Wang, Linxi Fan, Jiankai Sun, Ruohan Zhang, Li~Fei-Fei, Danfei Xu, Yuke Zhu, and Anima Anandkumar.
\newblock Mimicplay: Long-horizon imitation learning by watching human play.
\newblock In \emph{Conference on Robot Learning (CoRL)}, 2023.

\bibitem[Xiao et~al.(2022)Xiao, Radosavovic, Darrell, and Malik]{xiao2022masked}
Tete Xiao, Ilija Radosavovic, Trevor Darrell, and Jitendra Malik.
\newblock Masked visual pre-training for motor control.
\newblock \emph{arXiv}, 2022.

\bibitem[Xiong et~al.(2021)Xiong, Li, Chen, Bharadhwaj, Sinha, and Garg]{xiong2021learning}
Haoyu Xiong, Quanzhou Li, Yun-Chun Chen, Homanga Bharadhwaj, Samarth Sinha, and Animesh Garg.
\newblock Learning by watching: Physical imitation of manipulation skills from human videos.
\newblock In \emph{International Conference on Intelligent Robots and Systems (IROS)}, 2021.

\bibitem[Xu et~al.(2023)Xu, Xu, Chi, Veloso, and Song]{xu2023xskill}
Mengda Xu, Zhenjia Xu, Cheng Chi, Manuela Veloso, and Shuran Song.
\newblock Xskill: Cross embodiment skill discovery.
\newblock In \emph{Conference on Robot Learning}, pages 3536--3555. PMLR, 2023.

\bibitem[Ye et~al.(2024)Ye, Jang, Jeon, Joo, Yang, Peng, Mandlekar, Tan, Chao, Lin, et~al.]{ye2024latent}
Seonghyeon Ye, Joel Jang, Byeongguk Jeon, Sejune Joo, Jianwei Yang, Baolin Peng, Ajay Mandlekar, Reuben Tan, Yu-Wei Chao, Bill~Yuchen Lin, et~al.
\newblock Latent action pretraining from videos.
\newblock \emph{arXiv preprint arXiv:2410.11758}, 2024.

\bibitem[Zawalski et~al.(2024)Zawalski, Chen, Pertsch, Mees, Finn, and Levine]{zawalski2024robotic}
Micha{\l} Zawalski, William Chen, Karl Pertsch, Oier Mees, Chelsea Finn, and Sergey Levine.
\newblock Robotic control via embodied chain-of-thought reasoning.
\newblock In \emph{Conference on Robot Learning (CoRL)}, 2024.

\bibitem[Zhai et~al.(2023)Zhai, Mustafa, Kolesnikov, and Beyer]{zhai2023sigmoid}
Xiaohua Zhai, Basil Mustafa, Alexander Kolesnikov, and Lucas Beyer.
\newblock Sigmoid loss for language image pre-training.
\newblock In \emph{Proceedings of the IEEE/CVF International Conference on Computer Vision}, pages 11975--11986, 2023.

\end{thebibliography}
